\documentclass[letterpaper, 10 pt, conference]{ieeeconf}

\IEEEoverridecommandlockouts                              

\usepackage{amsmath,amssymb,amsfonts}
\usepackage{multirow}
\usepackage{graphicx}
\usepackage{rotating}
\usepackage{multicol}
\usepackage{amsmath}
\usepackage{float}
\usepackage{array}
\usepackage{caption}
\usepackage{subcaption}
\usepackage{siunitx}
\usepackage{float}
\usepackage{xcolor}
\usepackage{multirow}
\usepackage{graphicx}
\usepackage{color, soul, colortbl}
\usepackage{nicematrix}
\usepackage{colortbl} 
\usepackage{hyperref}
\usepackage[utf8]{inputenc}

\DeclareUnicodeCharacter{2061}{}

\begin{document}

\title{\LARGE Conditional Variational Auto Encoder Based Dynamic Motion for Multi-task Imitation Learning}

\author{Binzhao Xu, Muhayy Ud Din, Irfan Hussain$^{\star}$%
\thanks{Binzhao Xu, Muhayy Ud Din  and Irfan Hussain are with Khalifa University Center for Autonomous Robotic Systems (KUCARS), Khalifa University of Science and Technology, Abu Dhabi, UAE.}
\thanks{$^{\star}$ Corresponding author: irfan.hussain@ku.ac.ae.}
}

\maketitle
\begin{abstract} 
The dynamic motion primitive-based (DMP) method is an effective method of learning from demonstrations. However, most of the current DMP-based methods focus on learning one task with one module. Although, some deep learning-based frameworks can learn to multi-task at the same time. However, those methods require a large number of training data and have limited generalization of the learned behavior to the untrained state. In this paper, we propose a framework that combines the advantages of the traditional DMP-based method and conditional variational auto-encoder (CVAE). The encoder and decoder are made of a dynamic system and deep neural network. Deep neural networks are used to generate torque conditioned on the task ID. Then, this torque is used to create the desired trajectory in the dynamic system based on the final state. In this way, the generated tractory can adjust to the new goal position. We also propose a finetune method to guarantee the via-point constraint. Our model is trained on the handwriting number dataset and can be used to solve robotic tasks -- reaching and pushing directly. The proposed model is validated in the simulation environment. The results show that after training on the handwriting number dataset, it achieves a 100\% success rate on pushing and reaching tasks.

\end{abstract}
\IEEEoverridecommandlockouts


%
\IEEEpeerreviewmaketitle

\section{Introduction}

Learning from demonstration(LfD) is an effective way to transfer desired skills to the robot. By observing and imitating a human or another agent's actions, the robot can learn to generate the desired behaviors.  
There are two main challenges of imitation learning. First, \textit{A large amount of demonstrations:} The robot needs to observe enough demonstrations to learn the desired behavior precisely. Second, \textit{The generalization of the learned behavior:} In the environment of a new task, the robot needs to adapt the learned behavior.

Among various LfD techniques, Dynamic Motion Primitive(DMP) \cite{DMP2013} based methods have achieved great success in handling the above-stated challenges. In contrast to exploration-based methods, such as deep reinforcement learning \cite{RL_re, goecks2020}, DMP-based methods require less training data and time. They enable the direct extraction of global task-related information from complete trajectories rather than learning from trial and error.
Furthermore, unlike fundamental imitation learning, such as behavior cloning \cite{zhang_deep}, which simply replicates expert behavior through supervised learning, DMP-based methods can be conditioned on task parameters and adapted to new situations based on various task parameters.

Currently, most of the motion primitive methods focus on a single task. 
For instance, Ijspeert et al.\cite{DMP2013} used a second-order dynamic system to generate desired trajectories. By defining primitive force, complex trajectories can be generated by combining these forces with different weights. However, this method is designed to learn from one demonstration and does not consider the relationship between different degrees of freedom (DOF) of the demonstration. Paraschos et al.\cite{PoMp2013} is the first study to use a probabilistic model to learn motion primitive from multiple demonstrations. More recent works \cite{Frank_2022} and \cite{przystupa2023deep} consider constraints and higher dimensions of observation space. Some recent works demonstrated the effectiveness of combining VAE with DMP in learning from demonstration. For instance in ~\cite{chen2016, chen2015}, VAE works as a decoder, and the DMP is trained in the low dimensional latent space of VAE. The study proposed in~\cite{noseworthy20a} is similar to our work. They use DMP to create a dataset with different task conditions. Then the CVAE is trained and conditioned on the task parameters. They also proposed an Adversarially Enforcing Independence(AEI) method to disentangle latent space and task parameters to reduce the generalization error for the new task. However, it mainly focuses on a single task. Since they still use a deep neural network-based CVAE, there exist generalization errors on untrained data.

\begin{figure}[t]
    \centering
    \includegraphics[width=.95\columnwidth]{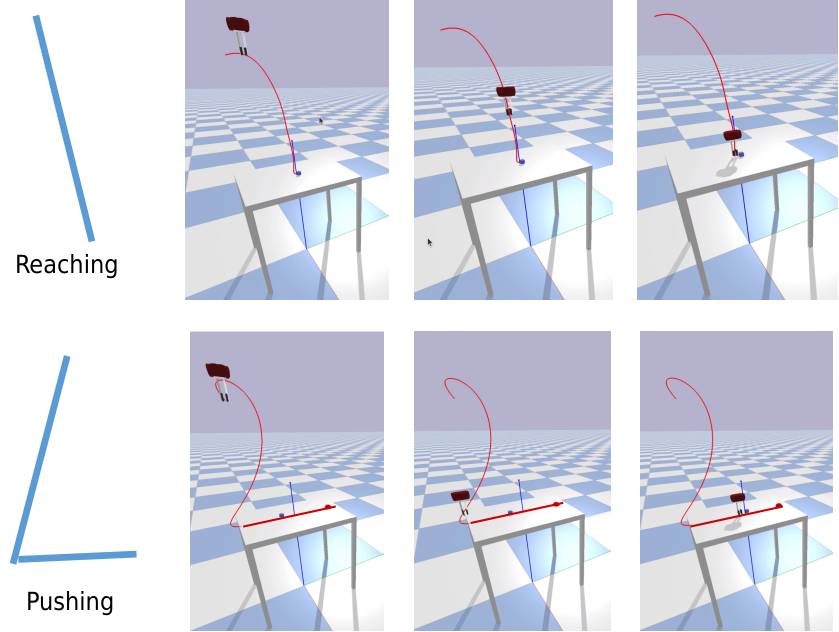}
    \caption{Mulit-task imitation learning. Different tasks have different shapes of trajectories. The trajectory for reaching the task looks like a line, while the trajectory for the pushing task is L shape.} 
    \label{fig:platform}
\end{figure}

In This study, we propose a variational auto-encoder(VAE) based dynamic motion technique to learn multi-task at the same time. Inspired by the works of DMP\cite{DMP2013}, and VAE\cite{Kingma_2019}. 
We employ a dynamic system to generate desired trajectories, and the VAE as a generative model to create the desired force. 
Our method consists of two stages: \textit{training stage} and \textit{generation stage}.  
During the training stage, we collected several normalized demonstrations for different tasks. The desired force is derived from an inverse dynamic system using these demonstrations. Then, a conditional variational auto-encoder (CVAE) model is trained to learn the distribution of the different forces.
In the generation stage, a force is sampled from the Decoder model based on the corresponding task class. This force is then input into the dynamic system to generate a normalized trajectory. We define an error between the normalized trajectory and the via-point(including the endpoint). Finally, the decoder and the scaler parameters are updated by minimizing the error.

The main contributions of our work are given below:
\begin{itemize}
\item \textit{Multi-task learning with few demonstrations:} We introduce a novel CVAE-based dynamic motion method capable of simultaneously learning multiple tasks with few demonstrations.
\item \textit{High-precision trajectory generation:} Our method takes advantage of dynamic motion and CVAE. The generated trajectory satisfied intial and final state automatically with the help of a dynamic motion system.  
\item \textit{Efficient trajectory generation:} The trajectory can be adjusted for the new task condition in a few iterations by fine-tuning the decoder and scaler parameters.  
\end{itemize}
\section{Problem Formulation}\label{sec:p-formulation} 

\subsection{Priliminiries}
This section will provide a brief introduction about the essential background of \textit{dynamic motion primitive} and \textit{conditional variational auto-encoder}.
\subsubsection{Dynamic Motion Primitive}
Dynamic Motion Primitive(DMP) is a method to generate desired trajectories by a second-order dynamic system. In general, this method has the advantage of guaranteeing the convergence of the trajectory to the goal state, and the scaleability of the trajectory. The dynamic system can be defined as:
\begin{equation}
    \tau \ddot{y} = \alpha(\beta(g-y)- \dot{y}) + f
\label{eq:dmp}
\end{equation}
Where: $\tau$ is the time constant, $y$ is the state in the trajectory, $g$ is the goal state, $\tau$ is the time scale, $\alpha$ and $\beta$ are the parameters of the dynamic system, $f$ is the force term.
This dynamic system defines the goal state as an attractor, guaranteeing the trajectory convergence to the goal state. The force $f$ is calculated as a weighted sum of the Gaussian basis functions. Similar to~\cite{DMP2013}, the force term can be defined as in Equation \ref{eq:force}.
\begin{equation}
    f = \frac{\sum_{i=1}^N w_i \psi_i(x)}{\sum_{i=1}^N \psi_i(x)}x(g-y_0)
\label{eq:force}
\end{equation}
Where, $w_i$ is the weight of the $i$th Gaussian basis function, $\psi_i(x)$ represents the $i$th Gaussian basis function, $x$ is the time factor decrease from 1 to 0, $y_0$ is the initial state of the trajectory.

From Equation \ref{eq:force}, we can observe that the desired force is proportional to the difference between the goal state and inital state. When we need to generate a trajectory to reach a new goal state, we just need to scale the force term to reach the desired goal state.

\subsubsection{Conditional Variational Auto-encoder} 
Variational Auto-encoder(VAE) is a generative model that can learn the distribution of the data set in the latent space. Conditional Variational Auto-encoder(CVAE) is a variant of VAE. The CVAE model is trained by maximizing the evidence lower bound(ELBO) of the data set conditioned on the task parameters. The ELBO of CVAE can be defined as:
\begin{equation}
\begin{aligned}   
    \mathcal{L}(\theta, \phi; x, c) = \mathbb{E}_{q_\phi(z|x, c)}[\log p_\theta(x|z, c)] - \\ D_{KL}(q_\phi(z|x, c)||p(z|c))
\label{eq:elbo}
\end{aligned}
\end{equation}
Where, $c$ is the task parameters, $\theta$, and $\phi$ are the parameters of the decoder and encoder, $x$ is the input data. $z$ is the latent variable, $q_\phi(z|x, c)$ represents the encoder, whereas, $p_\theta(x|z, c)$ represents the decoder, the term $p(z|c)$ is the prior distribution of the latent variable conditioned on the task parameters.

CVAE has been widely used in computer vision as a generative model to create images. Inspired by the work of Doersch et al. \cite{doersch2021tutorial}, the CVAE model is trained on the mnist data set where numbers are generated in pixels. In robot applications, different tasks usually have different shapes of trajectories. For example, the reaching task has a trajectory that looks like a straight line, while the trajectory of the grasping task looks like a v shape. The robot goes to the object's position and then lifts it to a certain height. It is a straightforward idea to generate different shapes of trajectories by conditioning the types of tasks.
\subsection{Problem Modeling}
In this section, we model the problem of multi-task imitation learning with CVAE-based DMP from the demonstration. The demonstration dataset $D$ is made with $m$ kinds of task, and each task has $n$ trajectories represented as $T$. The dataset $D$ can be defined as:
\begin{equation}
    D = \{ D_i = (T^i_k)_{k=1}^n\}_{i=1}^m
\end{equation}
Where: $D_i$ is the $i$th task, $T_k$ is the $k$th trajectory of the $i$th task. 

Each trajectory in $T$ consists of $L$ state points, each of which is $d$-dimensional. Additionally, the trajectory incorporates user-defined task parameters, including the task ID and via-points. The trajectories can be defined as:

\begin{equation}
    T_k = \{(x_l)_{l=1}^L, i, C \}  
\end{equation}
Where: $x_l \in \mathbb{R}^d$ is the $l$th state point of the trajectory, $i$ is the task id, $C$ is the set of the via-points. 

The goal of our method is to learn a trajectory generator $G$ from data set $D$. The new desired trajectories can be generated from $G$ with the given task parameters.

\section{Proposed Approach}

\begin{figure*}[t]
    \centering
    \includegraphics[width=2\columnwidth]{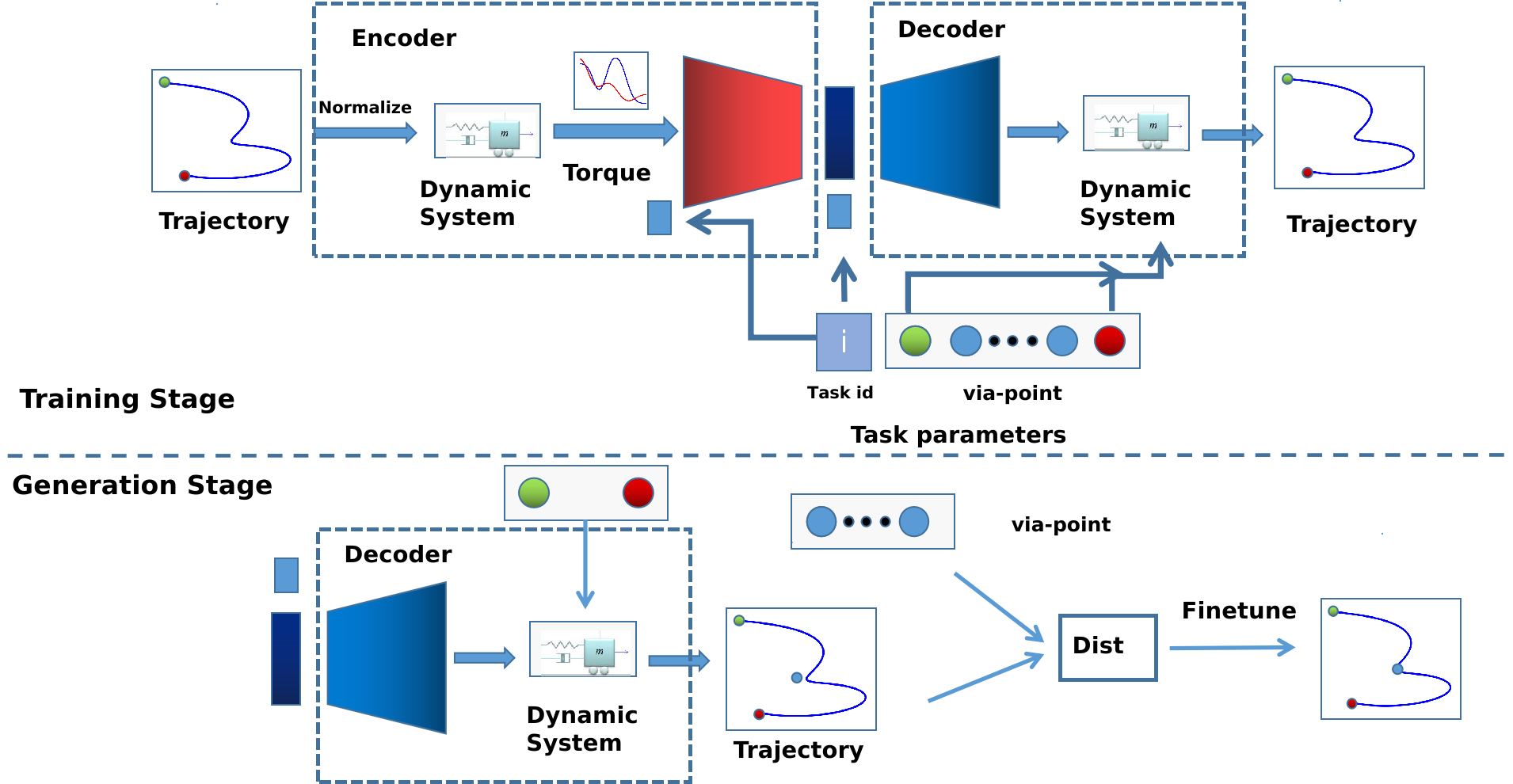}
    \caption{The main process of our method. Our method contains two stages -- the training stage and the generation stage. At the training stage, a CVAE structure is optimized and conditioned on task ID and via points. The encoder and decoder are made up of a second-order dynamic system and a deep neural network. At the generation stage, the new trajectory is generated by sampling from latent space and task parameters.} 
\label{fig:frame}
\end{figure*}

This section provides the details of our proposed CVAE-based DMP approach. Our method takes advantage of the flexibility of deep CVAE and the convergence and scaleability of DMP. The main process of the proposed approach is shown in Figure \ref{fig:frame}.

The proposed method consists of two stages: the training stage and the generating stage. During the training stage, we focus primarily on the trajectory's initial point, endpoint, and overall shape. The model comprises two main components: an encoder and a decoder. Each of these components incorporates a second-order dynamic system and a deep neural network.
In the encoder component, the trajectories are normalized to [0, 1]. Then, these normalized trajectories are input into the dynamic system to get the desired force with a 1D-CNN-based deep neural network.
In the decoder component, the desired force is sampled from the latent space of the CVAE. This force is then scaled based on the error between the initial and final states. Finally, the reconstructed trajectory is obtained from the dynamic system.

During the generation stage, the decoder module works as a trajectory generator based on the task ID, initial state, and goal state. Then the parameters of the decoder are tuned by the distance between the generated trajectory and the via-points. 

In the rest of this section, we provide the details of each step. To help the reader understand our method easily, we use the number handwriting dataset to illustrate the process.

\subsection{Date Augmentation}
\begin{figure}
    \centering
    \includegraphics[width=1\columnwidth]{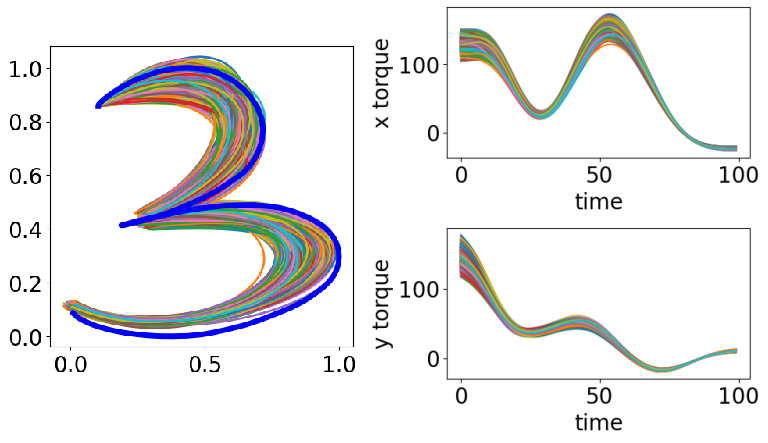}
    \caption{Data augmentation, the blue line represents the original trajectory, and the colored lines are the trajectories generated using data augmentation. right-side figures represent the forces of these trajectories.}
\label{fig:aug}
\end{figure}

Training the deep CVAE module requires hundreds of data. However, it is time-consuming to build a such large dataset from human demonstrations. To solve this issue, we use data augmentation to generate new trajectories. Firstly, the trajectories are normalized to $[0, 1]$. Then, we use the traditional DMP method to calculate the desired force as in Equation \ref{eq:force}. We add a Gaussian noise to the weight of each Gaussian basis function to obtain a new force. Finally, these forces are input into the dynamic system to generate a new trajectory. The data augmentation result is shown in Figure \ref{fig:aug}. In Figure \ref{fig:aug}, the blue line is the original trajectory, and the color lines are the new trajectories generated by the data augmentation. The right figures are the force of these trajectories by adding Gaussian noise to the weight in the Euqation \ref{eq:aug_force}.

\begin{equation}
    f = \frac{\sum_{i=1}^N w_i(1+k\varepsilon) \psi_i(x)}{\sum_{i=1}^N \psi_i(x)}x(g-y_0)
\label{eq:aug_force}
\end{equation}
Where: $k$ is the scale factor, k is 0.1 in this paper. $\varepsilon$ is the Gaussian noise.

\subsection{Training Stage}

At the training stage, we train a deep CVAE model to learn the distribution of the latent space of the trajectories, and able to reconstruct them. The modified deep CVAE model contains two parts: which are the encoder and the decoder. The encoder is made up of a dynamic system $Dy$, and a force encoder $En$ which is made up of 1D-CNN and fully connected layers. The decoder has a similar structure -- a force decoder $De$ and the same dynamic system $Dy$. 

Given a trajectory $T_k$, we obtain the force $F$ by the inverse dynamic system: $F = Dy^{-1}(T_k)$. The latent value $z$ is calculated by the force encoder as $z = En(F|i)$. $i$ is the task ID. This structure is inspired by the work of dosrsh \cite{doersch2021tutorial}, where they use a 2D CNN network as an encoder. In the number handwriting, $c$ is from 0 to 9. The force encoder is made up of 1D CNN and fully connected layers. 

In the decoder module, there is a force decoder trained to reconstruct the force $F'$ from the latent variable $z$. $F' = De(Z|i)$. Finally, the reconstructed trajectory is obtained by the dynamic system: 
$T' = Dy(k*F'|x_0, g)$. 
Where: $x_0$ is the initial state of the trajectory, and $g$ is the goal state in equation \ref{eq:dmp}. $k$ is the scale parameter of the force term. The loss function of the training stage can be defined as:

\begin{equation}
    \begin{aligned}   
        \mathcal{L}(\theta, \phi; x, c, x_0, g) = \sum_{D}(T_k - Dy(s*F'|x_0, g))^2 - \\D_{KL}(En_\phi(z|x, c)||De(z|c))
    \end{aligned}
\label{con_loss}
\end{equation}\\
Where $T_k$ is the $k$th trajectory in the data set $D$. $x_0$ and $g$ are the initial state and goal state of the trajectory. $s$ is the scale parameter of the force term that needs to be optimized.

Since the dynamic system is deterministic, and known, Equation \ref{con_loss} can be solved in an engineering way. We can first train to optimize the force encoder and decoder with $F$ and $F'$ by equation \ref{eq:elbo}, then update the scale parameters $s$ by minimizing the error between the end state and the goal state.

\begin{equation}
    s_i = s_i^0 * \frac{g_i - T_i^{end}}{T_i^{end} - T_i^{0}}
\label{eq:scale}
\end{equation} \\
Where $s_i^0$ is the initial value of the scale parameter, $g_i$ is the goal state of the $i$th trajectory, $T_i^{end}$ and $T_i^{0}$ are the end state and the initial state of the $i$th trajectory.

\subsection{Generating Stage}
At the trajectory generation stage, firstly, we generate a trajectory that satisfies the initial state, final state, and the shape of the task requirement. Then we adjust the trajectory to meet the via-points by finetuning the decoder parameters. The whole process is shown in the generation state in Figure \ref{fig:frame}.

\subsubsection{Generate with initial shape}
We generate a trajectory by sampling the latent variable $z$ from latent space. Force $F$ is created by the force decoder $De$ with the sampled $z$ and task id $i$. Then, the force term is input to the dynamic system with the equation \ref{eq:dmp} to generate a normalized trajectory $T$. The normalized trajectory can be scaled to the trajectory that satisfies the initial and final state by the scale of the force term with parameter $s$ in Equation \ref{eq:scale}. 

\begin{figure}[H]
    \centering
    \includegraphics[width=1\columnwidth]{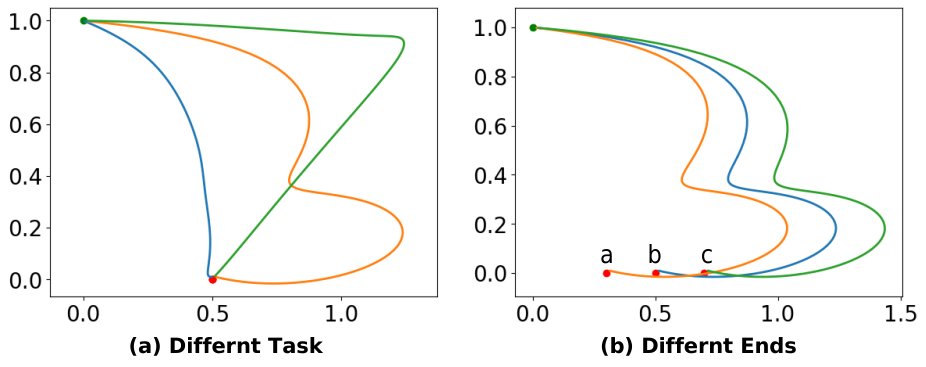}
    \caption{Trajectory Generation for different tasks and endpoints. Figure(a), all the trajectories begin at $[0,1 ]$, and end at $[1, 0]$. Figure(b), trajectory ends at different points.}
\label{fig:dif}
\end{figure}

As shown in Figure \ref{fig:dif}, we can generate trajectories with different shapes that satisfy the initial and final state constraints. In the Fig.~\ref{fig:dif}-(a), the trajectories are begin at point $[0, 1]$ and ends at point $[0.5, 1]$. These trajectories are generated by the task id -- 1, 3, 7. The trajectory can go from the beginning point to the endpoint with the corresponding shape.

Since the final trajectories are generated by the dynamic system, our model inherits the scaleability of DMP. In Figure \ref{fig:dif}(b), we generate 3 trajectories with the same task id -- number 3. The trajectories are begin at point $[0, 1]$ and ends at points a:$[0.3, 0]$, b:$[0.5, 0.5]$, c:$[0.8, 0]$. Once the torque is generated by the torque decoder module, we just need to scale the torque using Equation \ref{eq:scale} to reach the desired goal state.

\subsubsection{Via point Constrain} 
To finish a task successfully, the trajectory should not only satisfy the initial and final state but also pass through the required via points. For instance, in the robotic pick and place task, the manipulator should go to the object position first, then grasp and move the object to a final position. 

\begin{figure}[H]
    \centering
    \includegraphics[width=1\columnwidth]{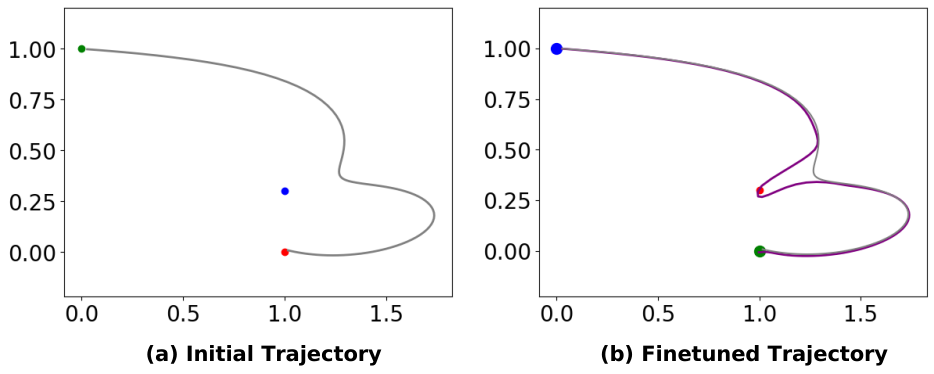}
    \caption{Finetune for via point constrain. Figure(a) is the initial trajectory. Figure(b) is the trajectory with via point constrain.}
\label{fig:tune}
\end{figure}

In this section, we finetune the decoder parameters to meet the via points. The loss function is defined in Equation \ref{tune_loss}. The loss function is made up of three terms. The first term is the error between the re-generated trajectory and the initial trajectory. This loss aims to keep the shapes of the re-generated trajectory. The second term is to limit the offset of the endpoint in the new trajectory. The final term is to minimize the distance between the via-points and the re-generated trajectory.

\begin{equation}
    \begin{aligned}   
        \mathcal{L}(C) = p_1* \sum_{T}(x'_i - x_i)^2 /L + p_2*(x'_{end} - g)^2 + \\ p_3 *\sum_{C}(dist(T', C_k) 
    \end{aligned}
\label{tune_loss}    
\end{equation}\\
Where: $C$ is the set of the via-points. $T$ is the inital trajectory, $T'$ is the trajectory generated after finetuning. $x_i$ is the $i$th point in the trajectory $T$. $g$ is the goal state of the new task. $dist(T', C_k)$ is the minimal distance between the via-point and the trajectory.

The force decoder $De$ contains three layers that fully connect the network. In this task, we find that freeze parameters of the first two layers can give the best fine-tunning result. In Figure \ref{fig:tune}(a) is the trajectory for the drawing number 3. Figure \ref{fig:tune}(a) is the initial trajectory generated by the force decode module. The blue point is the required via-point. The finetuned trajectory is shown in Figure \ref{fig:tune}(b) red line. We can see that the finetuned trajectory can pass through the via-point while keeping the shape similar to the initial trajectory.

\section{Experiment}\label{sec:exp}
We use the handwriting task and robot task in the gym environment to show the effectiveness of our method. To evaluate our method in a systematic way, we use the following evaluation metrics:
\begin{itemize}
    \item Via-point error. The minimal distance between the via-point and the trajectory.
    \item Final state error. If no object in the environment, the final state error is the distance between the goal state and the endpoint of the trajectory. Otherwise, the error is the distance between the goal state and the object position. 
    \item Success rate/Trajectory shape. For the robotic manipulation task in the gym environment, this is the success rate of the task. For the handwriting task, which does not have a specific success state definition, we compare the shape of the desired trajectory with the generated.
\end{itemize}


\subsection{Number Handwriting}
\begin{figure}
    \centering
    \includegraphics[width=1\columnwidth]{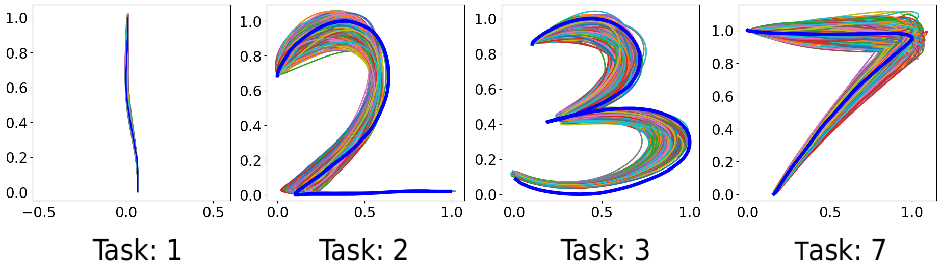}
    \caption{Data augmentation for 4 tasks.}
\label{fig:num_task}
\end{figure}

In this paper, we do not consider the rhythmic movements which have the same start and end state. For the handwriting data, numbers $1, 2, 3, 7$ are selected. The original data set and the one with data augmentation are shown in Figure \ref{fig:num_task}. We select deep CVAE as the baseline. 

\subsubsection{Final State Error}
\begin{figure}[H]
    \centering
    \includegraphics[width=1\columnwidth]{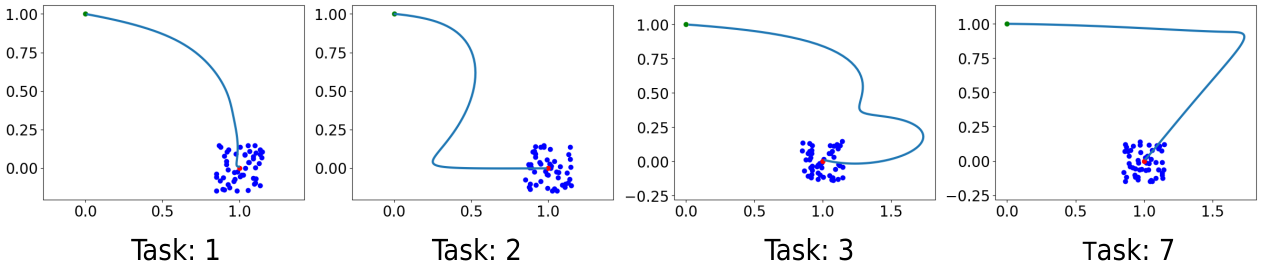}
    \caption{Test error for different endpoints.}
\label{fig:num_ends}
\end{figure}

For each task, the trajectory begins at the $[0, 1]$ and ends at a point in the point $[1, 0] + 0.3 * r$, where $r$ is standard Gaussian noise. The final state and the end point of each task are shown in Figure \ref{fig:num_ends}. We randomly select 50 endpoints for each task. The training time of our method at the training stage is around 12 seconds. The final state error for each task is shown in Table \ref{error}. For each, the error of the initial trajectory is below 0.02 for the normalized trajectory. After finetuning, the final state error is below 0.008. Among them, task 1 -- writing number 1 has the lowest error. This is because the shape of the trajectory is simple.

\subsubsection{Via-point Error}
\begin{figure}[H]
    \centering
    \includegraphics[width=1\columnwidth]{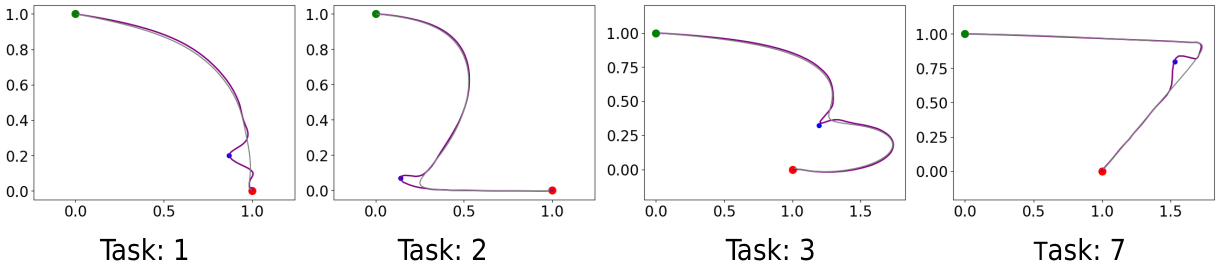}
    \caption{Test error for via-point.}
\label{fig:num_via}
\end{figure}

In this section, we test the adjustability of our methods in the unseen via-point. For each task, the via-point is selected as the point offset the mid of the trajectory by -0.2 from the x-axis.  Figure \ref{fig:num_via}, shows that our method can adjust the trajectory to the new via-point while still keeping the shape of the trajectory. The retraining time for finetuning each task takes around 2.2 seconds. From Table \ref{error}, we can see that the via-point error is below 0.005 for each task. This means that if the workspace of the robot is around 1m, the via-point error is below 5 mm which is acceptable for most of the robotic tasks.

\subsubsection{Trajectory Shape}
\begin{figure}
    \centering
    \includegraphics[width=1\columnwidth]{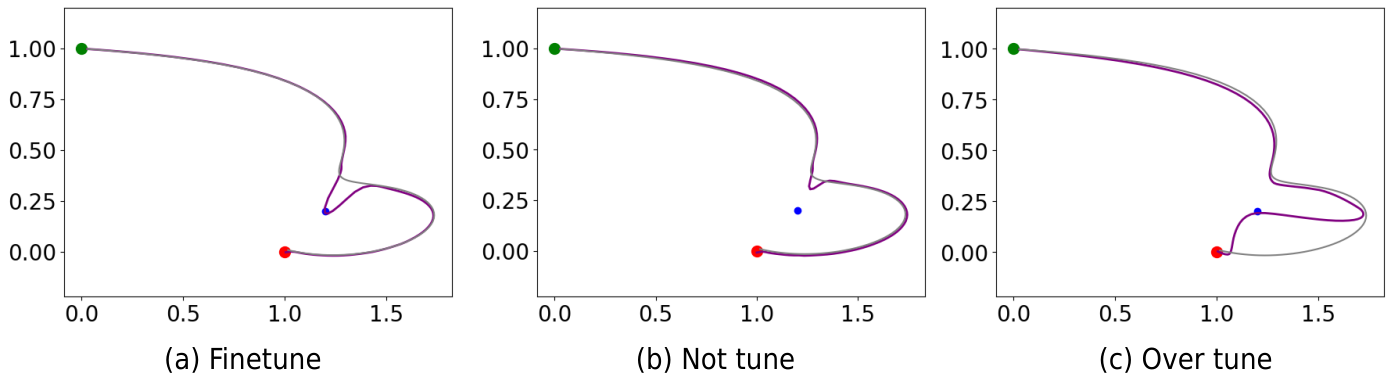}
    \caption{Different finetune paramter. (a) $p_1 = 0.6, p_2 = 0.2, p_3 = 0.2$. (b) $p_1 = 0.9, p_2 = 0.05, p_3 = 0.05$. (c) $p_1 = 0.05, p_2 = 0.05, p_3 = 0.9$.}
\label{fig:diff_tune}
\end{figure}

We also tested the shape of the generated trajectory. From Equation \ref{tune_loss}, there are three contradictory terms. The first term is to keep the shape of the newly generated trajectory to be close to the initial trajectory. The second term is to limit the offset of the endpoint in the new trajectory. The final term is to minimize the distance between the via-points and the new trajectory. If $p_1$ is too big, the new trajectory can not blend to the via-point, as shown in Figure\ref{fig:diff_tune}(b). If $p_3$ is too big, the trajectory may shrink to the via-point Figure\ref{fig:diff_tune}(c), and the shape of the trajectory will be changed. In this study, we set $p_1 = 0.6, p_2 = 0.2, p_3 = 0.2$. These parameters can give a balance of via-point tuning and shape-keeping. From Table \ref{error}, the shape error of the trajectory is less than 0.015.

\begin{table}[H]
    \centering
    \begin{tabular}{ccccc}
    \hline
            & DM\_CVAE  & \multicolumn{3}{c}{T\_DM\_CVAE} \\ \hline
    Task id & End Error & End Error  & Via-point & Shape  \\ \hline
    Task  1 & 0.0102    & 0.0024     & 0.0027    & 0.0142 \\
    Task 2  & 0.0153    & 0.0098     & 0.0024    & 0.0133 \\
    Task 3  & 0.0179    & 0.0017     & 0.0045    & 0.0104 \\
    Task 7  & 0.0184    & 0.0071     & 0.0012    & 0.0116 \\ \hline
    \end{tabular}
    \caption{\label{error} Error of our methods.}
\end{table}

\subsection{Robotic Task}
We also apply our trajectory generator to the robotic manipulation task. We use the model trained on the number handwriting dataset directly. It is interesting to see that the trained model has a success rate of 1 on the reaching task and object push task. We create a simulation environment for a robotic manipulator with Baxter's hand in Pybullet \cite{coumans2021}. 

These environments are created similar to Fetch environments in the gym\cite{gymnasium}. Since the handwriting model is trained on the 2D dataset, we constrain the manipulator's movement in these environments 
\subsubsection{Reaching Task}
\begin{figure}[H]
    \centering
    \includegraphics[width=1\columnwidth]{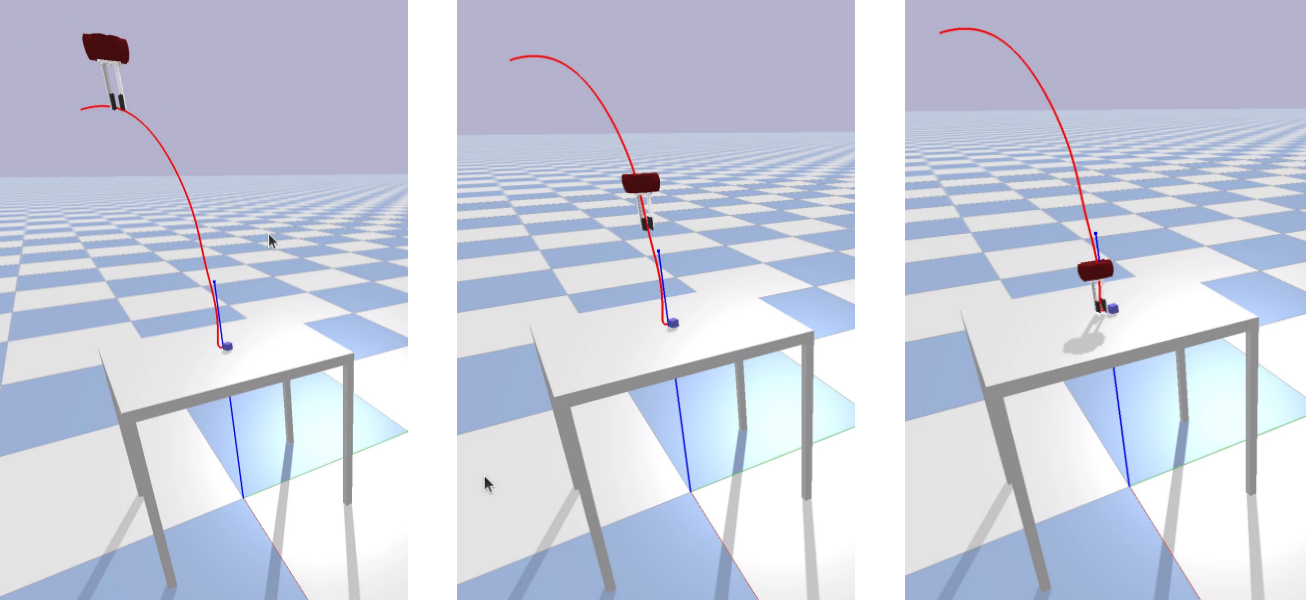}
    \caption{Reaching task.}
\label{fig:reaching}
\end{figure}

The requirement of this task is to touch the blue cube on the table. We select this environment as a representation of the task which does not require the via-point constraint. For the reaching task, the trajectory looks similar to number one, therefore we select the task parameter as $1, x0, g$, where $1$ is the task id, and $x0$, and $g$ are the beginning point and end point of the trajectory. From Figure \ref{fig:reaching}, the gripper touches the object with a shape that looks similar to the number 1.

\subsubsection{Pushing Task}
\begin{figure}[H]
    \centering
    \includegraphics[width=1\columnwidth]{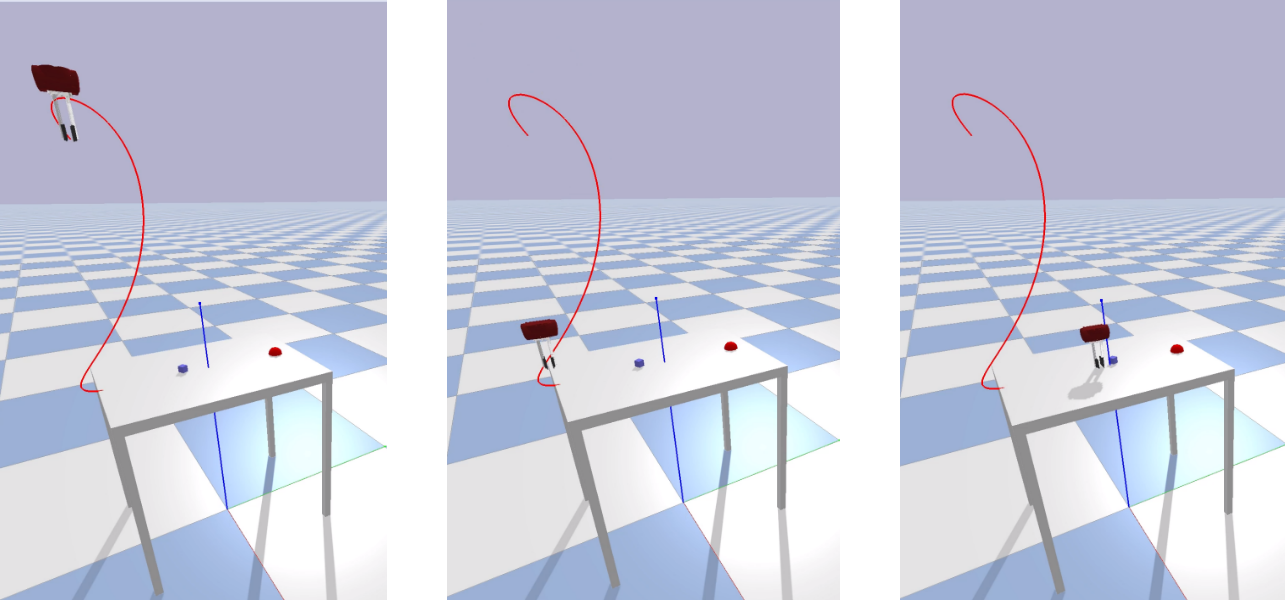}
    \caption{Pushing task.}
\label{fig:pushing}
\end{figure}

This environment is used to test our method on the robotic task which has via-point constrain. In this environment, the robot should go to a cube position and then push it to a goal position (red dot in Figure \ref{fig:pushing}). We select the trajectory of shape $2$ to finish this task. As shown in Figure \ref{fig:pushing}, the manipulator goes to a point behind the object, then approaches it and pushes it to the desired position. 

\section{Conclusions}\label{sec:conclusion}
In this paper, we propose a CVAE-based DMP method that inherits the flexibility of the deep CVAE, and the scalability property of DMP. It can learn to generate trajectories for different tasks in one module. Similar to DMP, it can adapt to different goal states. The effectiveness of this method has varied on the handwriting data set. The trained method can applied to the robotic task -- pushing and reaching. However, in the robotic manipulation task, the correct trajectory type has to be selected by ourselves based on our prior experiment. Although the trajectory generated by our module can achieve a high success rate, it is not optimal. In the later work, we are trying to create a module trained by more kinds of shape trajectories. The best type of trajectory for the task should also be selected automatically.

\bibliographystyle{ieeetr}
\bibliography{references}

\smallskip

\end{document}